# Survey of NLU Benchmarks Diagnosing Linguistic Phenomena: Why not Standardize Diagnostics Benchmarks?

Khloud AL Jallad[1], Nada Ghneim[2], Ghaida Rebdawi[1]
Khloud.aljallad@hiast.edu.sy, n-ghneim@aiu.edu.sy, ghaida.rebdawi@hiast.edu.sy
[1] Informatics Department, Higher Institute of Applied Sciences and Technology, Damascus, Syria.
[2] Faculty of Informatics and Communication, Arab International University, Daraa, Syria.

## Abstract

Natural Language Understanding (NLU) is a basic task in Natural Language Processing (NLP). The evaluation of NLU capabilities has become a trending research topic that attracts researchers in the last few years, resulting in the development of numerous benchmarks. These benchmarks include various tasks and datasets in order to evaluate the results of pretrained models via public leaderboards. Notably, several benchmarks contain diagnostics datasets designed for investigation and fine-grained error analysis across a wide range of linguistic phenomena. This survey provides a comprehensive review of available English, Arabic, and Multilingual NLU benchmarks, with a particular emphasis on their diagnostics datasets and the linguistic phenomena they covered. We present a detailed comparison and analysis of these benchmarks, highlighting their strengths and limitations in evaluating NLU tasks and providing in-depth error analysis. When highlighting the gaps in the state-of-the-art, we noted that there is no naming convention for macro and micro categories or even a standard set of linguistic phenomena that should be covered. Consequently, we formulated a research question regarding the evaluation metrics of the evaluation diagnostics benchmarks: "Why do not we have an evaluation standard for the NLU evaluation diagnostics benchmarks?" similar to *ISO standard* in industry. We conducted a deep analysis and comparisons of the covered linguistic phenomena in order to support experts in building a global hierarchy for linguistic phenomena in future. We think that having evaluation metrics for diagnostics evaluation could be valuable to gain more insights when comparing the results of the studied models on different diagnostics benchmarks.

## Keywords

Natural Language Understanding (NLU), Textual Entailment, Textual Inference, Natural Language Inference (NLI), Diagnostics Datasets, Arabic Language Understanding Evaluation (ALUE), General Language Understanding Evaluation (GLUE), Framework for Computational Semantics (FraCaS).

## 1. Introduction

Natural Language Understanding (NLU) is a fundamental core of Natural Language Processing (NLP), yet its performance suffers from weaknesses in handling the complexities of human languages, ranging from syntactic ambiguity to high-level reasoning difficulties. In the recent years, several NLU benchmarks were published allowing comparative evaluation of pretrained models via public leaderboards. However, having a high score in NLU tasks is not a such meaningful detailed evaluation, as it does not allow NLU designers to know the effective weaknesses and strengths of a studied model. Addressing these weaknesses can be done through error analysis across diverse linguistic phenomena and in-depth investigation using diagnostics datasets. Diagnostics dataset allows in-depth investigation through error analysis across diverse linguistic phenomena. By analyzing model performance on these specific phenomena, NLU designers can better understand their model's generalization behavior, gain valuable insights and develop targeted strategies to overcome NLU weaknesses, ultimately leading to more robust and generalizable models.

In this paper, we conducted a survey of available English, Arabic, and Multilingual NLU benchmarks in which we compared covered task clusters and datasets, with a particular emphasis on diagnostics datasets. We studied the macro and micro linguistic phenomena in these diagnostics benchmarks. When highlighting the gaps in the state-of-the-art (SoTA), we noted that there is no naming convention for macro or micro categories or even a standard set of linguistic phenomena that should be covered. Therefore, we posed a research question regarding the evaluation metrics of the evaluation diagnostics benchmarks: "Why do not we have an evaluation standard for the NLU evaluation diagnostics benchmarks?". We conducted a deep analysis and comparisons of the linguistic phenomena in order to support NLP/ linguistics experts in building a global hierarchy in future. Although determining the necessary "amount of data required to produce relevant performance measures" remains an open problem [1], we have just made initial statistics on macro/micro categories counts, distributions across linguistic phenomena and diagnostics data sizes, but it still lacks important evaluation criteria, such as: language difficulty level, diversity of covered topics, variety of covered NLP levels in samples (Pragmatic, Discourse, …), etc. We hope that this open question will help moving forward into future researches regarding standardizing diagnostics benchmarks.

This paper is organized as follows; Section 1 is the introduction; state-of-the-art is shown in section 2. Section 3 contains statistics, analysis and comparisons. Finally, discussions are in section 4.

The contributions of this paper are:
1- Survey of available diagnostics datasets with a focus on their studied linguistic phenomena.
2- When highlighting the gaps in SoTA, we noted that there is no naming convention for macro and micro categories or even a standard set of linguistic phenomena that should be covered. Consequently, we posed a research question regarding the evaluation metrics of the evaluation diagnostics benchmarks: "why do not we have an evaluation standard for the NLU evaluation diagnostics benchmarks?". We suggested to build a global hierarchy for linguistic phenomena under supervision of linguistics experts.

## 2. Related Works

### 2.1. Natural Language Inference

To the best of our knowledge, the first research work that defines natural language inference was in 1973 at Stanford university [2] that converts all available knowledge to a canonical template form and endeavors to create chains of non-deductive inferences from the unknowns to the possible referents. Its method of selecting among possible chains of inferences is consistent with the overall principle of "semantic preference" used to set up the original meaning representation, of which these anaphoric inference procedures are a manipulation.

Textual Entailment (TE) was originally proposed by Dagan and Glickman in 2004 [3] as a generic paradigm for applied semantic inference, and subsequently established through the series of benchmarks known as the *PASCAL Recognizing Textual Entailment (RTE) Challenges* from 2005 up to 2011 [4], [5], [6], [7], [8], [9], [10], [11], [12]. *Recognizing Textual Entailment (RTE)* is the task of detecting whether a sentence meaning can be entailed from another sentence. The first sentence is called Text (T) or Premise (P), the second sentence is called Hypothesis (H), RTE task is to determine if a reader reads T or P would conclude H or not. It was called 2-way RTE. Later another type of RTE arrived in 2008, the 3-way-RTE [13], also known as *Natural Language Inference (NLI)*, which is the task of determining relation between sentences pairs to more detailed entailment (entailment, neutral or contradiction). An Example of entailments relations types is shown in Table 1.

| Text (Premise) | I enjoy research work. | Class |
|---|---|---|
| **Hypothesis 1** | I hate research. | Contradiction |
| **Hypothesis 2** | I like ice cream. | Neutral |
| **Hypothesis 3** | I like reading research papers. | Entailment |

*Table 1: Entailment Pairs Examples*

NLI task could be a valuable evaluation method for NLU. The reason behind that is NLI ability to encompass complex language understanding skills, from resolving syntactic ambiguity to high-level reasoning, in a straightforward simple canonical form: recognizing when the meaning of a text snippet is contained in the meaning of a second piece of text. This simple abstraction of an extremely complex evaluation, made NLI a crucial component of any technological advancement program[14].

Several NLI and RTE datasets were constructed in many languages. Also, several models were published by many universities and companies. In this section, we will cover few NLI works in English and Arabic only.

As for English language, the most famous available datasets are: WNLI [15], MNLI [16], SNLI [17]. Also, GLUE proposed RTE dataset [18] which is a combination of RTE1 [12], RTE2 [11], RTE3 [10], and RTE5 [9]. Some state-of-the-art models in English are: PaLM 540B[19], Vega_v2_6B[20], RoBERTa[21],

SemBERT[22], XLNET[23], AlexaTM_20B[24], BloombergGPT and GPT-NeoX and BLOOM_176B [25], UnitedSynT5[26], ByT5[27], Rethinking Coupling[28], mGPT[29], DeBERTa[30], SpanBERT[31], SqueezeBERT[32], DistilBERT[33].

As for Arabic language, some available datasets are: ArbTEDS corpus [34], ArNLI dataset [35], ArEntil Dataset [36]. Furthermore, some multilingual datasets contain Arabic section such as SNLI dataset [37], XNLI dataset [38]. Some state-of-the-art models that contain Arabic language are: Facebook Bart Large MNLI [39], [40], Facebook RoBERTa-Large-MNLI [41], Microsoft Multilingual MiniLM [42], Microsoft XLM-RoBERTa [42], Microsoft DeBERTaV3 [43]. Moreover, Laurer et al. [44] tuned many versions of pretrained multilingual models such as: Moritzlaurer Ernie-M-Base-MNLI-XNLI, mDeBERTa-V3-Base XNLI-Multilingual-NLI-2mil7, Moritzlaurer-Multilingual-MiniLMv2-L6-MNLI-XNL, Moritzlaurer-Multilingual-MiniLMv2-L12-MNLI-XNLI.

## 2.2. State of the art Benchmarks

There are two main types of textual benchmarks in NLP:
- Natural Language Understanding (NLU) Benchmarks.
- Natural Language Generation (NLG) Benchmarks.

Few benchmarks were designed for evaluating both NLU and NLG tasks such as, GPTAraEval [45], but most of them were either for evaluating NLU or NLG. In this paper, we will focus on NLU Benchmarks, while mentioning the most important NLG Benchmarks without deeply discussing their detailed categories (see Table 2).

| CATEGORY | BENCHMARK | LANGUAGE | # TASK CLUSTERS | # DATASETS |
|---|---|---|---|---|
| MULTILINGUAL | CLSE[46] | 3 LANGUAGES | 1 | 1 |
|  | GEM$_{v1}$[47] | 18 LANGUAGES | 5 | 13 |
|  | GEM$_{v2}$[48] | 51 LANGUAGES | 9 | 40 |
|  | INDICNLG[49] | 11 LANGUAGES | 5 | 5 |
|  | MTG[50] | 5 LANGUAGES | 4 | 4 |
|  | INDONLG[51] | 3 LANGUAGES | 4 | 10 |
| X-SPECIFIC | DOLPHIN[52] | AR | 13 | 40 |
|  | ARBENCH[53] | AR | 1 | 5 |
|  | ARAOPUS-20[54] | AR | 1 | 1 |
|  | ARGEN[55] | AR | 7 | 13 |
|  | GPTARAEVAL[45] | AR | 13 | 23 |
|  | BANGLA-NLG[56] | BN | 6 | 7 |
|  | PHOMT[57] | VI | 1 | 1 |
|  | CUGE[58] | ZH | 8 | 9 |
|  | LOT[59] | ZH | 2 | 2 |
|  | BAHSA INDONISIA[60] | ID | 1 | 14 |
|  | GLGE [61] | EN | 4 | 8 |

*Table 2: NLG Benchmarks General Statistical Comparison*

Although there are several NLU benchmarks in SoTA, not all them contains diagnostics datasets. Table 3 shows a comparison between SoTA benchmarks regarding number of tasks, number of datasets, available languages and whether containing diagnostics dataset or not.

| CATEGORY | BENCHMARK | LANGUAGE | DIAGNOSTICS | # TASKS | # DATASETS |
|---|---|---|---|---|---|
| MULTI-LINGUAL | XGLUE[62] | 19 LANGUAGES | - | 3 | 11 |
| | XTREME[63] | 40 LANGUAGES | ✓ | 4 | 9 |
| | XTREME-R[64] | 50 LANGUAGES | ✓ | 4 | 10 |
| X-SPECIFIC | GLUE[18] | EN | ✓ | 5 | 11 |
| | SUPER GLUE[65] | | - | 5 | 10 |
| | ALUE[66] | AR | ✓ | 3 | 9 |
| | ARLUE[67] | | - | 4 | 42 |
| | ORCA[68] | | - | 7 | 60 |
| | ARABIC MMLU[69] | | - | 1 | 40 |
| | LARABENCH[70] | | - | 7 | 61 |
| | BABI[71] | EN, HI | - | 2 | 20 |
| | FLUE [72] | FR | - | 5 | 7 |
| | KORNLU[73] | KOR | - | 2 | 4 |
| | CLUE[74] | ZH | ✓ | 6 | 9 |
| | JGLUE[75] | JP | - | 3 | 6 |
| | INDONLU [76] | ID | - | 5 | 12 |
| | KNLU[77] | KOR | - | 8 | 8 |
| | UINAUIL[78] | IT | - | 6 | 6 |
| | SUPERGLUER[79] | DE | - | 3 | 29 |
| | VLUE[80] | VI | - | 5 | 5 |
| | INDICGLUE[81] | EN, HI | - | 11 | 11 |

Table 3: NLU Benchmarks General Statistical Comparison

### 2.3. Diagnostics and Linguistic Phenomena

Diagnostics dataset is not a test set for machine learning models performance, it is a specialized evaluation dataset that is used by humans to pinpoint specific areas where models struggle. Diagnostics dataset allows in-depth investigation through error analysis across diverse linguistic phenomena. By analyzing model performance on these specific phenomena, NLU designers can better understand their model's generalization behavior, gain valuable insights and develop targeted strategies to overcome NLU weaknesses. This could ultimately lead to more robust and generalizable models.

As for the structure of diagnostics dataset, each instance is a pair of sentences, labeled with NLI classes and tagged with the coarse-grained categories corresponding to the linguistic phenomena. Each of these coarse-grained categories has several fine-grained subcategories.

The first paper that proposed a linguistic phenomenon hierarchy framework for computational semantics was published by Stanford university in 1996. This paper introduced FraCaS[1] (Framework for Computational Semantics) framework [82]. FraCaS corpus consists of 346 units, so-called *problems*, each including 1-5 statements (premises), a "yes/no question", and a "yes/no" answer, where "yes" indicates an entailment, "no" a contradiction, and "don't know" a neutral case. The problem set has examples of a wide variety of *problems* in formal semantics: generalized quantifiers, negation, monotonicity, anaphora, ellipsis, comparatives, adjectives, temporal relations, and propositional attitudes. FraCas coarse-grained categories are shown in Figure 2 (a). FraCas fine-grained categories are shown in Figure 3. However, FraCas project did not include any evaluations, and there was no follow-up. The first study that used a part of the FraCaS suite was held by MacCartney and Manning in 2009 [83] where the FraCaS corpus was improved, represented as XML, and the questions were rephrased into declarative sentences, facilitating the automatic processing. While FraCas does indeed systematically cover a wide range of linguistic phenomena, the small size of the dataset limits the ability to meaningfully extrapolate from the results. Although determining the necessary "amount of data required to produce relevant performance measures" remains an open problem [1].

---
[1] https://nlp.stanford.edu/~wcmac/downloads/

Later, in 2010 Bentivogli et. al [84] proposed a methodology to follow when creating specialized TE datasets. They carried out a feasibility study applying their methodology to a sample of 90 pairs extracted from the RTE-5 dataset [4]. It was not a diagnostics dataset, but they discussed several linguistic phenomena. Their studied linguistic phenomena were 5 coarse-grained categories: Lexical, Lexical-Syntactic, Syntactic, Discourse, Reasoning (Figure 2 (b)). Each of them has its detailed fine-grained as follows (Figure 4):
- **Lexical:** identity, format, acronymy, demonymy, synonymy, semantic opposition, hyperonymy, geographical knowledge.
- **Lexical-syntactic:** transparent heads, nominalization/verbalization, causative, paraphrase.
- **Syntactic:** negation, modifier, argument realization, apposition, list, coordination, active/passive alternation.
- **Discourse:** coreference, apposition, zero anaphora, ellipsis, statements.
- **Reasoning:** apposition, modifiers, genitive, relative clause, elliptic expressions, meronymy, metonymy, membership/representativeness, reasoning on quantities, temporal and spatial reasoning, all the general inferences using background knowledge.

Researches were conducted to analyze specialized TE datasets, such as [85], and to design an evaluation specialized dataset, such as quantitative reasoning [86], [87], comprehensive reading [87], common sense reasoning [88], and translation representation analysis [89].

A study was held in 2022 to create a test suite for sub-clausal negation only, NaN-NLI [90], to facilitate fine-grained analysis of model performance and provide deeper understanding of the current NLI capabilities in terms of negation and quantification. NaN-NLI contains premise–hypothesis pairs, where the premise contains sub-clausal negation, and the hypothesis is constructed by making minimal modifications to the premise in order to reflect different possible interpretations. It covers the following negation types (*verbal vs. non-verbal, analytic* vs. *synthetic, clausal vs. sub-clausal*) with a focus on *sub-clausal* type. Moreover, it contains the following operations (*Indefinite quantifier change, Numerical quantifier change, Comparative quantifier change, Negator addition or deletion, Negator position change, Negator token change, Clause or sub-clause deletion, Focus particle change, Lexical change, Syntactic change*). The dataset was manually annotated in terms of negation types, negation constructions.

The first *Diagnostics dataset* GLUE (General Language Understanding Evaluation) benchmark [18] was introduced in 2018. GLUE contains 1104 pairs, categorized into 4 coarse-grained categories: Lexical Semantics, Predicate-Argument Structure, Logic, Knowledge and Common Sense (Figure 2 (c)). Each of them has its detailed fine-grained subcategories as follows (Figure 5).
- **Lexical Semantics**: Lexical Entailment; Morphological Negation; Factivity; Symmetry/Collectivity; Redundancy; Named Entities; Quantifiers.
- **Predicate-Argument Structure**: Syntactic Ambiguity; Prepositional Phrases; Core Arguments; Alternations; Ellipsis/Implicits; Anaphora/Coreference; Intersectivity; Restrictivity.
- **Logic**: Propositional Structure; Quantification; Monotonicity; Richer Logical Structure.
- **Knowledge And Common Sense**: World Knowledge; Common Sense.

As for CLUE (Chinese Language Understanding Evaluation) [74], the benchmark does not contain fine-grained categories. CLUE only contains 9 coarse-grained categories: *Anaphora, Argument Structure, Common Sense, Comparative, Double Negation, Lexical Semantics, Monotonicity, Negation, Time of event*. (Figure 2 (d)).

The first *Arabic Diagnostics dataset* ALUE (Arabic Language Understanding Evaluation) benchmark [33] was proposed in 2021. ALUE contains 1147 pairs, categorized into 5 coarse-grained categories: Lexical

Semantics, Predicate-Argument Structure, Logic, Quantification, Knowledge and Common Sense (Figure 2 (e)). Each of them has its detailed fine-grained subcategories (Figure 6).

- **Lexical Semantics**: Lexical Entailment; Morphological Negation; Symmetry/Collectivity; Redundancy; Named entities; Quantifiers
- **Predicate-Argument Structure**: Relative clauses; Prepositional Phrase; Alternation (Causative/Inchoative; Active/Passive; Nominalization; Datives; Topicalization); Anaphora/Coreference; Intersectivity; Restrictivity
- **Logic**: Negation; Double negation; Conjunction; Disjunction; Conditionals; Monotonicity (Upward entailment; Downward entailment; non-Monotone)
- **Quantification** Universal; Existential
- **Knowledge and Common Sense**: World Knowledge; Common Sense

Several benchmarks were published after ALUE for benchmarking in Arabic language [67] [68] [69] [70]. However, they did not contain new diagnostics datasets.

A multilingual benchmark (XTREME) (Cross-lingual TRansfer Evaluation of Multilingual Encoders) [63], created pseudo test sets by automatically translating the English test set to 40 other languages. Researchers in XTREME highlighted that these translations were noisy and should not be treated as ground truth. For that reason, we do not discuss its diagnostics linguistic phenomena. Same for XTREME-R.

Different structures were used to represent the categorization of linguistic phenomena into subcategories. FraCas used a problem set for each phenomenon. GLUE has the macro categories as columns names and their values are the used micro-categories if any, or null otherwise. As for ALUE, it has columns for macro-categories and columns for micro-categories having zeros and ones to indicate the presence of a phenomenon or not. Figure 1 shows few samples from FraCaS(a), GLUE(b) and ALUE(c) datasets, respectively.

```
fracas-012      answer: undef **
P1   Few great tenors are poor.
Q    Are there great tenors who are poor?
H    There are great tenors who are poor.
A    Not many

fracas-013      answer: yes
P1   Both leading tenors are excellent.
P2   Leading tenors who are excellent are indispensable.
Q    Are both leading tenors indispensable?
H    Both leading tenors are indispensable.

fracas-014      answer: no
P1   Neither leading tenor comes cheap.
P2   One of the leading tenors is Pavarotti.
Q    Is Pavarotti a leading tenor who comes cheap?
H    Pavarotti is a leading tenor who comes cheap.
```

(a): FraCas Diagnostics Samples

| Lexical Semantics | Predicate-Argument | Logic | Knowledge | Domain | Premise | Hypothesis | Label |
|---|---|---|---|---|---|---|---|
| Quantifiers | | Universal | | Artificial | Everyone has a set of principles to live by. | Someone has a set of principles to live by. | entailment |
| Quantifiers | | Universal | | Artificial | Someone has a set of principles to live by. | Everyone has a set of principles to live by. | neutral |
| Quantifiers | | Universal | | Artificial | Everyone has a set of principles to live by. | No one has a set of principles to live by. | contradiction |
| Quantifiers | | Universal | | Artificial | No one has a set of principles to live by. | Everyone has a set of principles to live by. | contradiction |
| Quantifiers | | Universal | | Artificial | Everyone has a set of principles to live by. | Susan doesn't have a set of principles to live by. | contradiction |

(b): GLUE Diagnostics Samples

| sentence1 | sentence2 | gold_label | Lexical Se | Predicate-A | Logic_coa | Knowledge_ | Symmetry/Coll | Factivity | Named e | Quantifiers | Lexi |
|---|---|---|---|---|---|---|---|---|---|---|---|
| أكلت بيتزا مع بعض الأصدقاء. | أكلت بيتزا. | entailment | 0 | 1 | 0 | 0 | 0 | 0 | 0 | 0 | 0 |
| أكلت بيتزا. | أكلت بيتزا مع بعض الأصدقاء. | entailment | 0 | 1 | 0 | 0 | 0 | 0 | 0 | 0 | 0 |
| أكلت بيتزا مع بعض الأصدقاء. | أكلت بعض الأصدقاء. | neutral | 0 | 1 | 0 | 0 | 0 | 0 | 0 | 0 | 0 |
| أكلت بعض الأصدقاء. | أكلت بيتزا مع بعض الأصدقاء. | neutral | 0 | 1 | 0 | 0 | 0 | 0 | 0 | 0 | 0 |
| أكلت بيتزا بالزيتون. | أكلت بيتزا. | entailment | 0 | 1 | 0 | 0 | 0 | 0 | 0 | 0 | 0 |
| أكلت بيتزا. | أكلت بيتزا بالزيتون. | neutral | 0 | 1 | 0 | 0 | 0 | 0 | 0 | 0 | 0 |
| أكلت بيتزا بالزيتون. | أكلت زيتون. | entailment | 0 | 1 | 0 | 0 | 0 | 0 | 0 | 0 | 0 |
| أكلت زيتون. | أكلت بيتزا بالزيتون. | neutral | 0 | 1 | 0 | 0 | 0 | 0 | 0 | 0 | 0 |

(c): ALUE Diagnostics Samples

*Figure 1: Samples from FraCas, GLUE, and ALUE Diagnostics Datasets.*

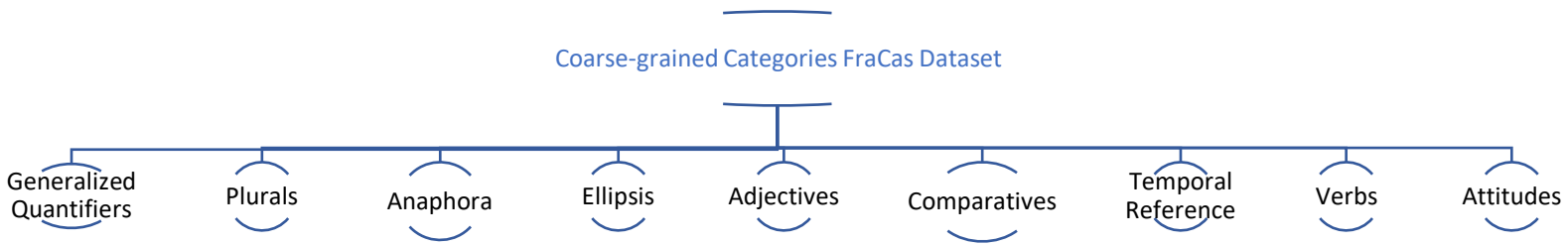

*(a) Coarse-grained Categories Fracas Dataset*

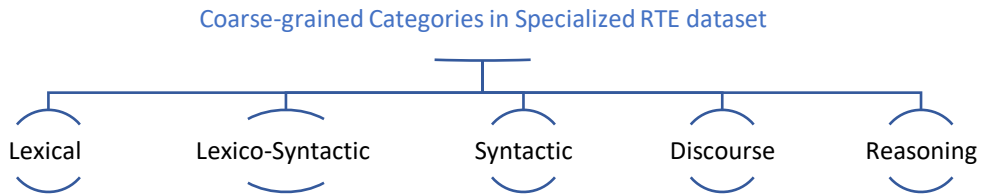

*(b) Coarse-grained Categories in Specialized RTE dataset*

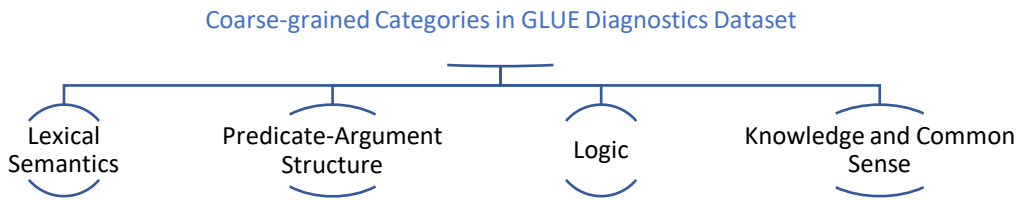

*(c) Coarse-grained Categories in GLUE Diagnostics Dataset*

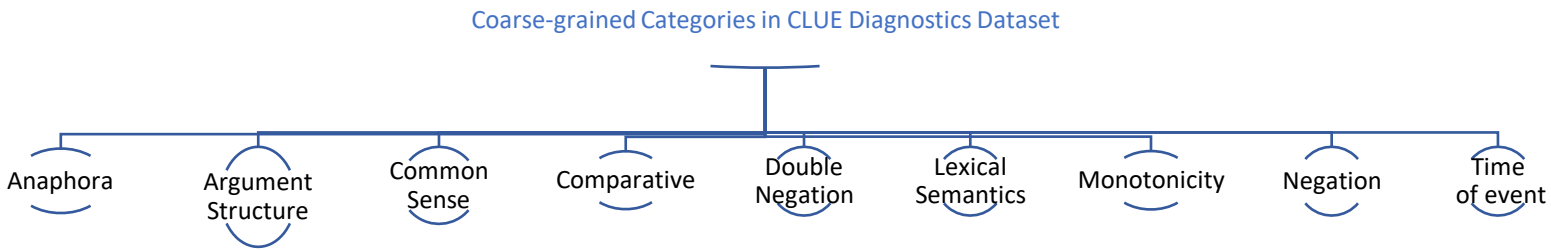

*(d) Coarse-grained Categories in CLUE Dataset*

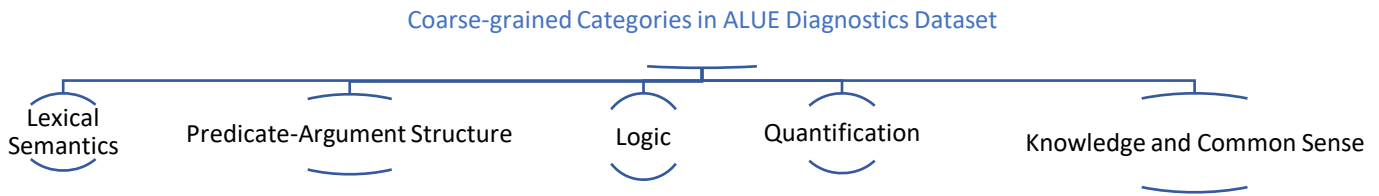

*(e) Coarse-grained Categories in ALUE Diagnostics Dataset*

*Figure 2: Coarse-grained Categories in SoTA*

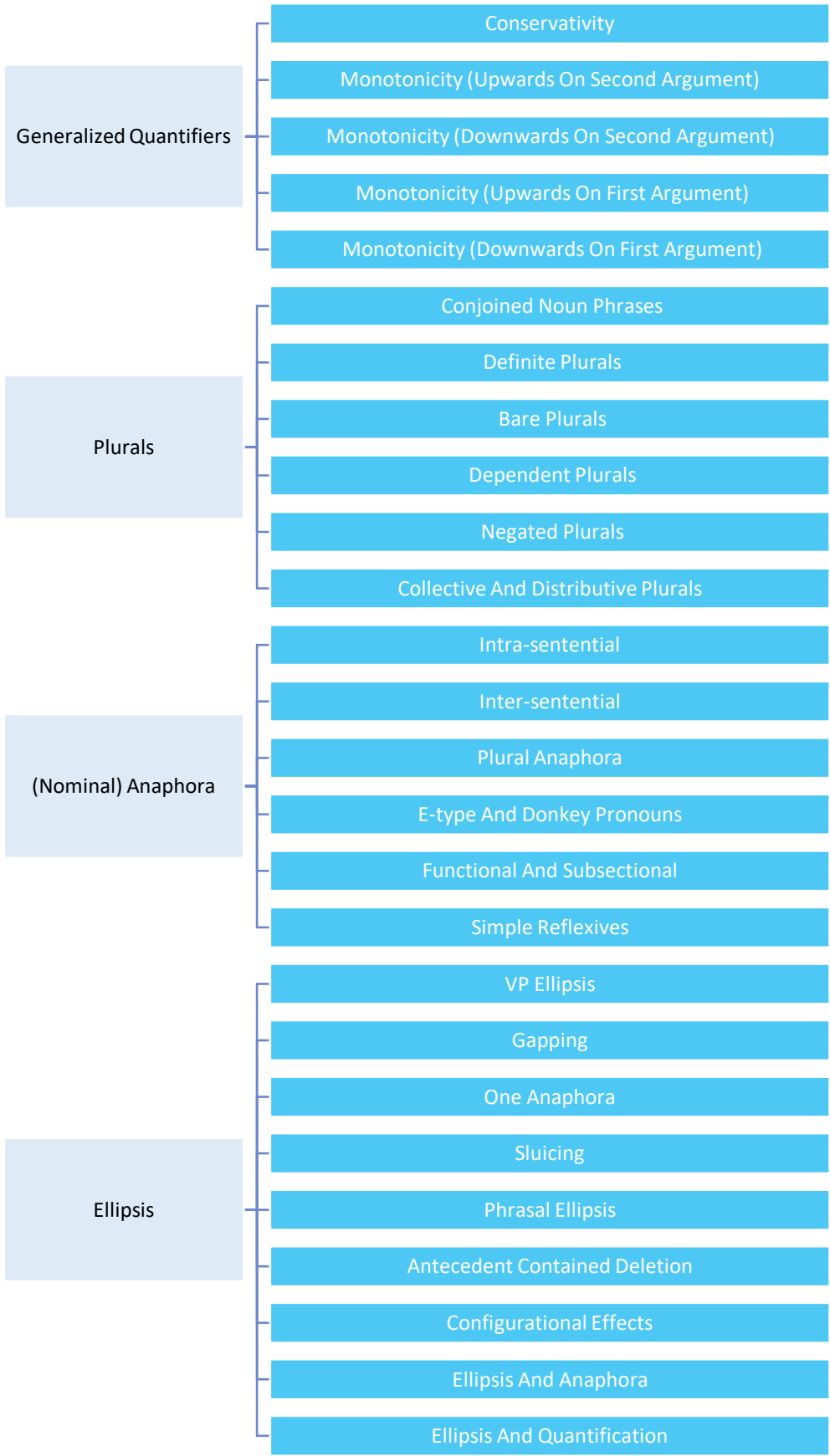

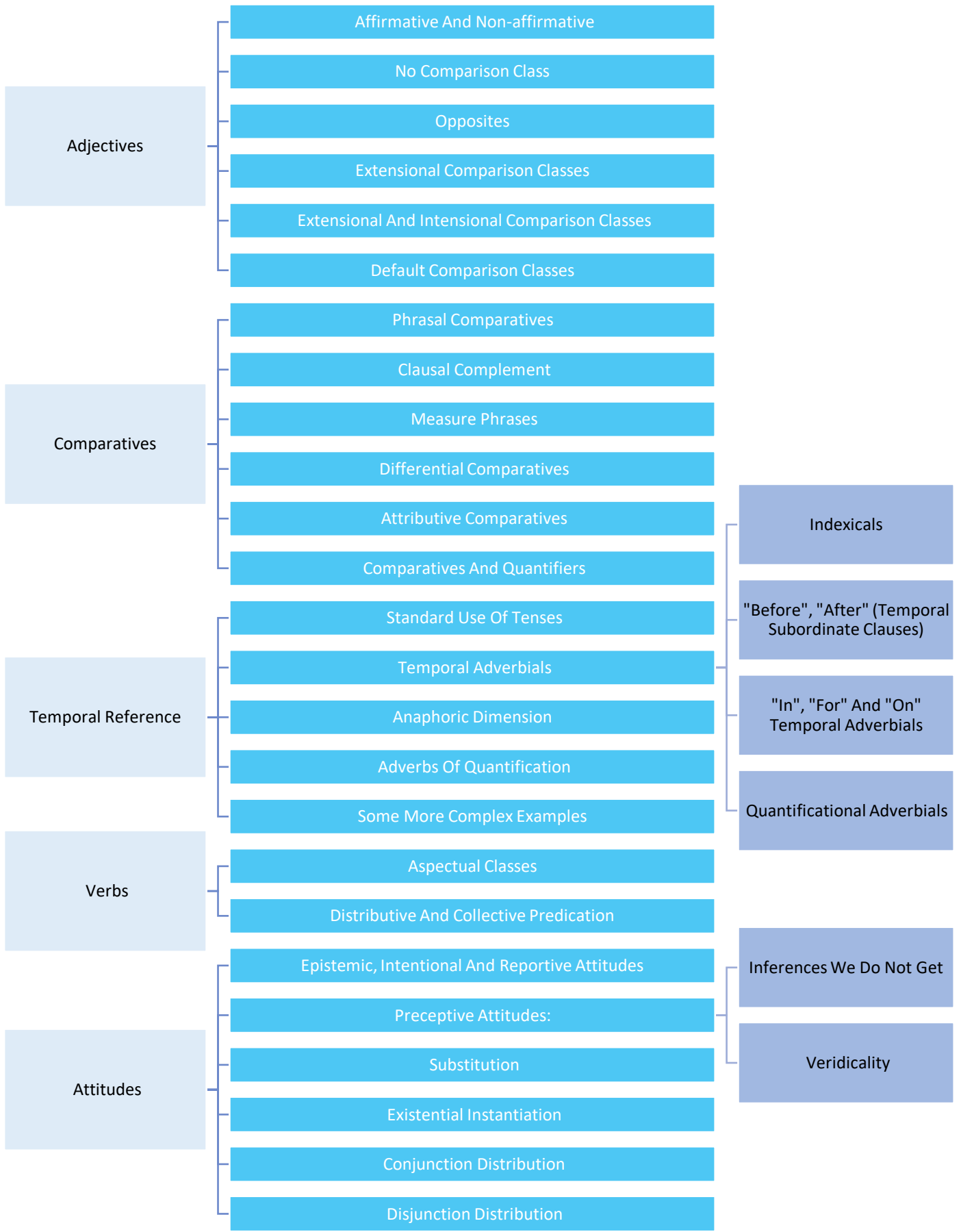

Figure 3: Fine-grained Categories for FraCas

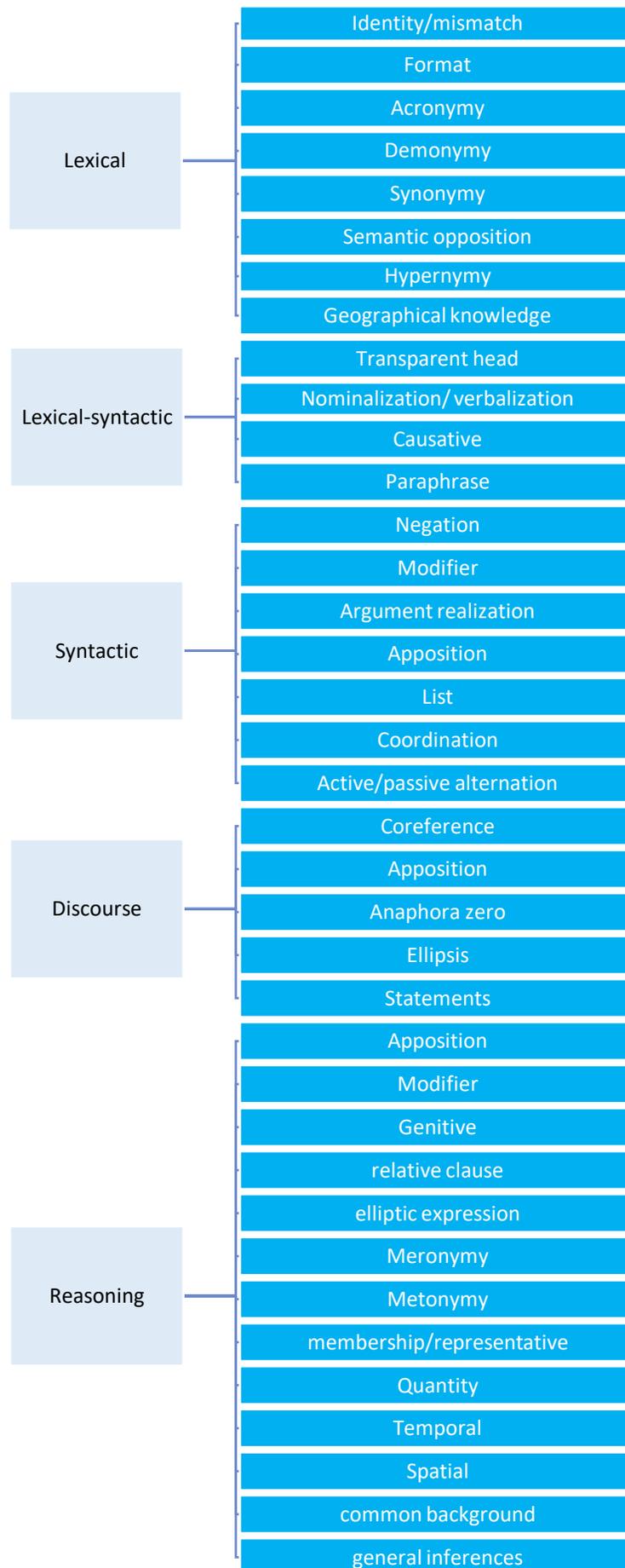

Figure 4: Fine-grained Categories for Specialized TE Dataset

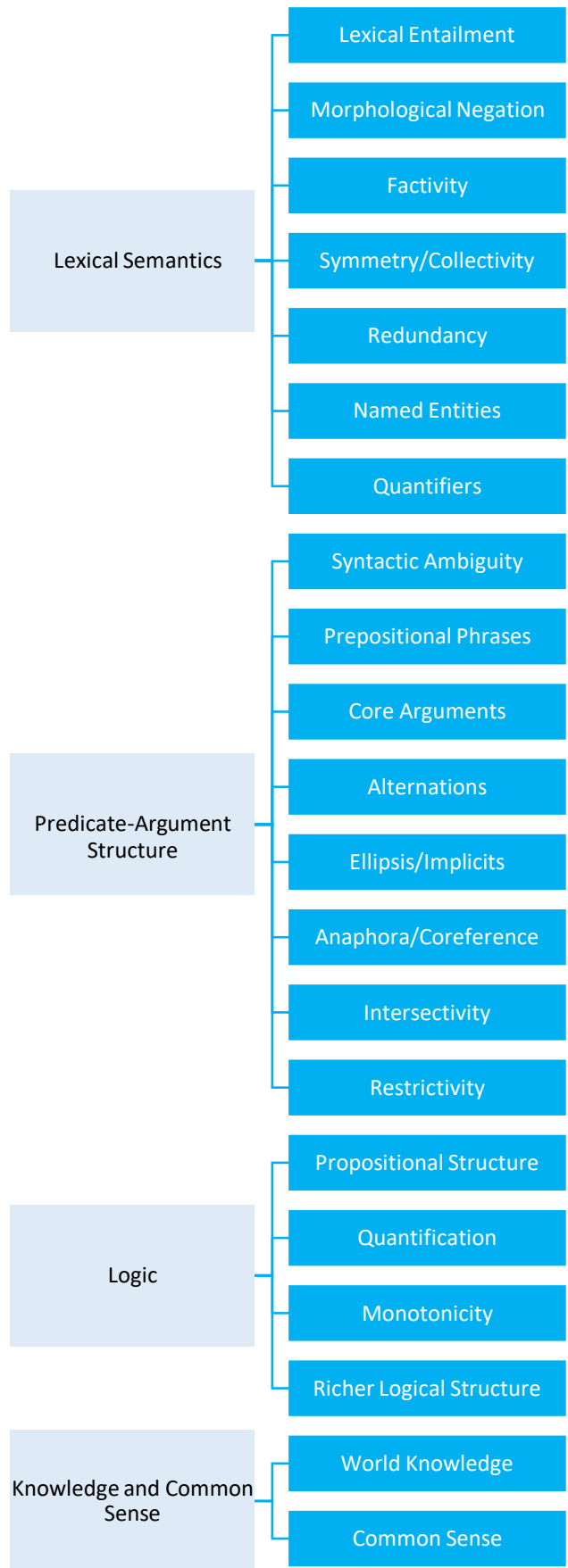

Figure 5: Fine-grained Categories in GLUE Diagnostics Dataset

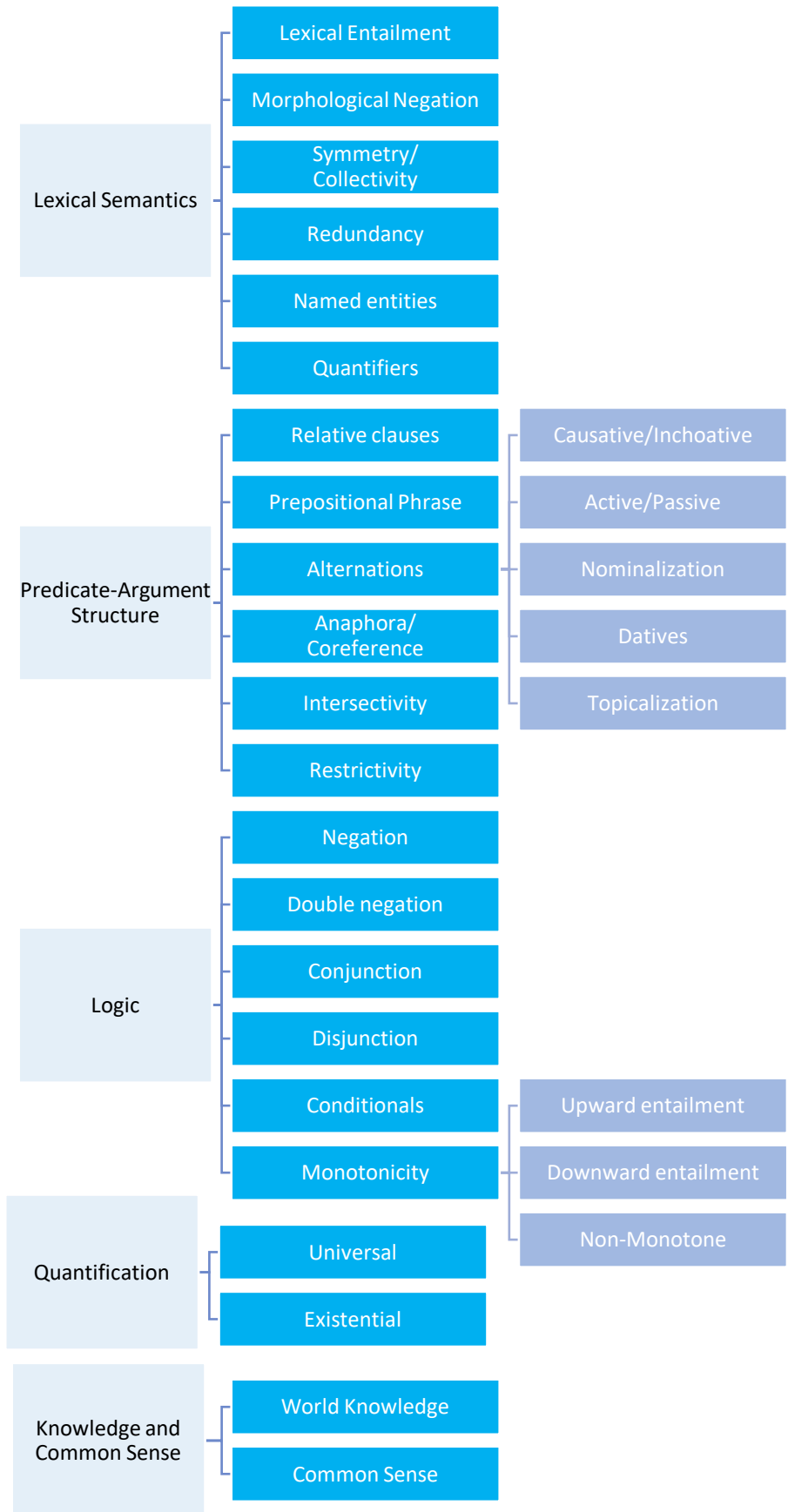

Figure 6: Fine-grained Categories in ALUE Diagnostics Dataset

## 3. Benchmarks STATISTICS & Analysis

We conducted basic analysis of SoTA diagnostics datasets. Figure 7 shows distribution of all macro-categories in SoTA.

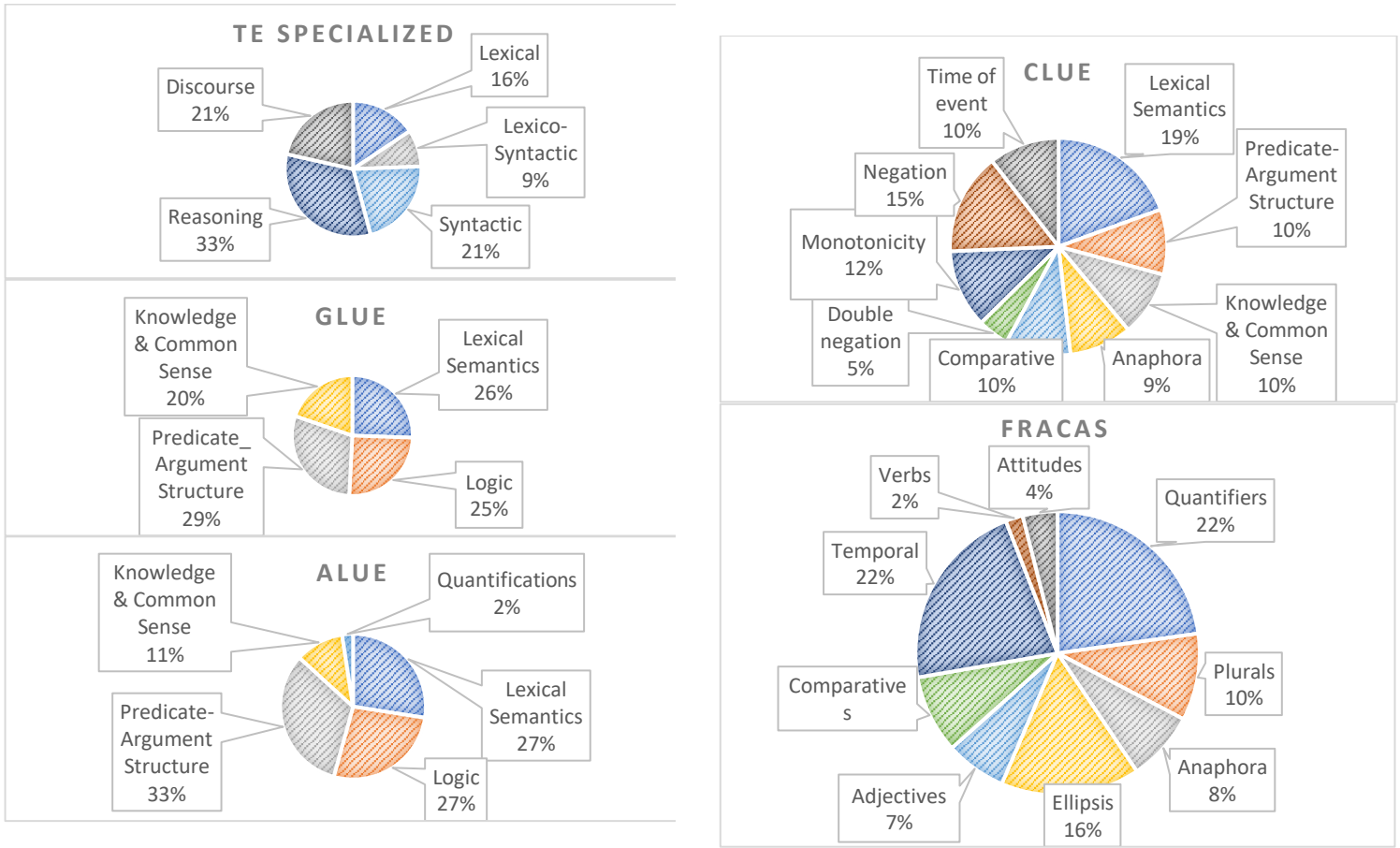

Figure 7: Distribution of Studied Diagnostics Datasets Macro-Categories Distribution

Table 4 shows benchmarks diagnostics samples statistics.

| Macro Category | FraCas | TE Specialized | GLUE | ALUE | CLUE |
|---|---|---|---|---|---|
| **Lexical** | - | 32 | - | - | - |
| **Lexical Semantics** | - | - | 368 | 477 | 100 |
| **Lexico-Syntactic** | - | 18 | - | - | - |
| **Syntactic** | - | 44 | - | - | - |
| **Reasoning** | - | 67 | - | - | - |
| **Discourse** | - | 44 | - | - | - |
| **Logic** | - | - | 364 | 466 | - |
| **Predicate-Argument Structure** | - | - | 424 | 564 | 50 |
| **Knowledge & Common Sense** | - | - | 284 | 188 | 50 |
| **Quantifications** | 80 | - | - | 42 | - |
| **Plurals** | 33 | - | - | - | - |
| **Anaphora** | 28 | - | - | - | 48 |
| **Ellipsis** | 55 | - | - | - | 50 |
| **Adjectives** | 23 | - | - | - | 24 |
| **Comparatives** | 31 | - | - | - | 60 |
| **Temporal** | 75 | - | - | - | 78 |
| **Verbs** | 8 | - | - | - | 54 |
| **Attitudes** | 13 | - | - | - | - |

Table 4: Benchmarks Diagnostics Samples Statistics

Figure 8 shows distribution of classes in studied diagnostics datasets, where we can see that most samples are of entailment class.

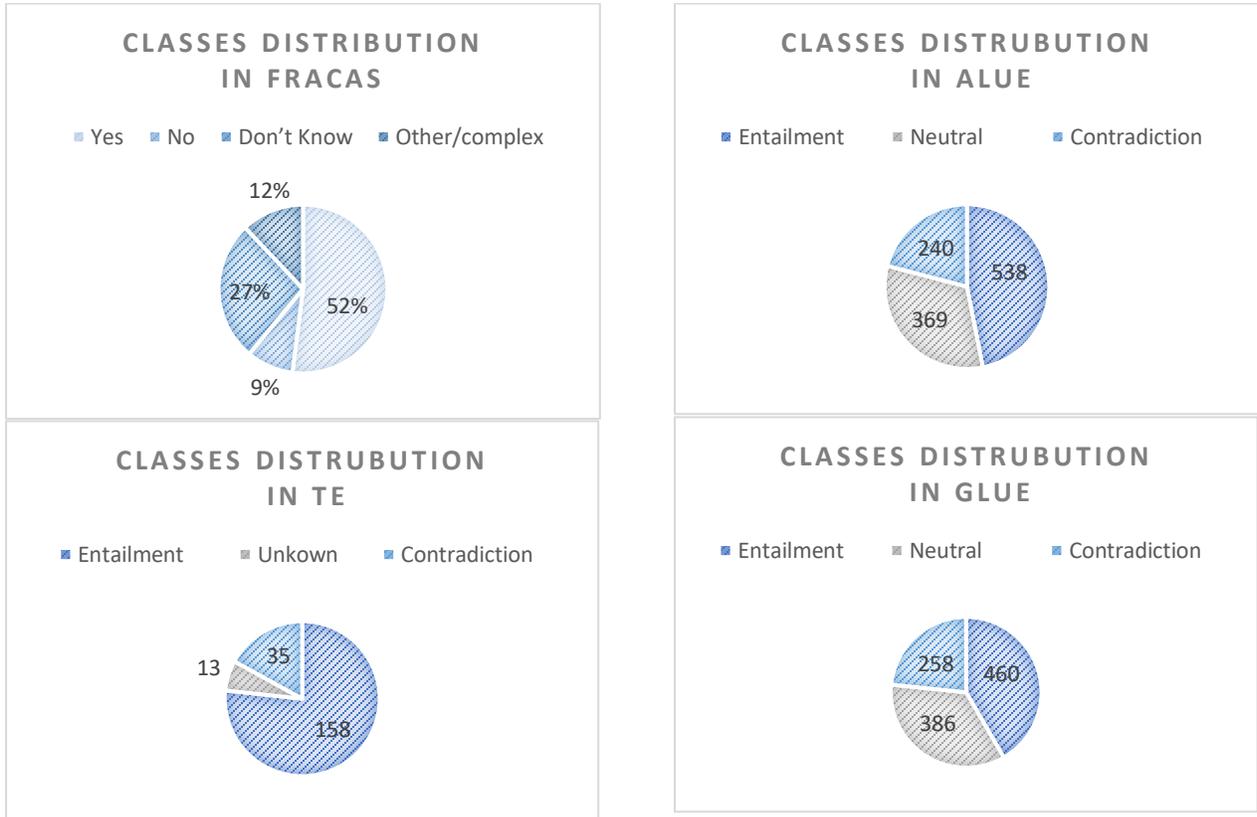

Figure 8: Distribution of Classes in Diagnostics Datasets

## 4. Discussion

In this section, we will discuss diagnostics phenomena based on general macro categories.

**Lexical, Lexico-Syntactic, Syntactic:**

Diagnostics datasets vary significantly in their categorization of syntactic and semantic phenomena. While GLUE and ALUE group syntactic phenomena into a single macro-category, i.e., *Predicate-Argument Structure*, older datasets like FraCas categorized each syntactic phenomenon as a separate macro-category. For instance, FraCas treats *Adjectives* as a macro-category, with micro-categories such as "Affirmative and Non-affirmative", "No Comparison Class", ... In contrast, GLUE and ALUE consider *Adjectives* as a micro-category within the broader *Predicate-Argument Structure* category.

Moreover, FraCas separates lexical semantics into multiple macro-categories such as, *Verbs* and *Attitudes*, whereas GLUE and ALUE group all lexical semantics into a single macro-category that includes *Lexical Entailment Morphological Negation, Factivity, Symmetry/Collectivity, Redundancy, Named Entities, and Quantifiers*.

On the other hand, the TE specialized dataset utilizes three macro-categories: LEXICAL, LEXICO-SYNTACTIC, SYNTACTIC, to address these phenomena.

**Ellipsis**

Ellipsis is a linguistic phenomenon where part of a sentence is omitted, as it is either understandable from the context or unnecessary as it is clear enough, thus the reader is able to implicitly fill the gap. Entailment examples can be constructed based on this phenomenon by explicitly filling the gap with the correct/incorrect referents. For example, the premise "Mark is a successful technical leader in our company, most employees can imagine no other leader" *entails* "Mark is a successful technical leader in our company, most employees can imagine no other leader than Mark" and

*contradicts with* "Mark is a successful technical leader in our company, most employees can imagine no other leader than themselves."

Ellipsis is a macro category in Fracas and a micro category denoted as *Ellipsis/Implicits* in ALUE and GLUE under the *Predicate-Argument Structure* macro category. GLUE argued that even Ellipsis is often regarded as a special case of anaphora, but they decided to split it out from explicit *anaphora*, which is often also regarded as a case of *coreference*.

**Logic & Reasoning**

Fracas has no logical category, while all other diagnostics include a logical macro category that is named *Logic or Reasoning*. Most logical categories include *Monotonicity* that will be covered in detail next section.
In ALUE and GLUE, the *Logic* macro-category includes *Negation, Double Negation, Conjunction, Disjunction, Conditionals*. In contrast, CLUE includes these categories as macro-categories rather than distinguishing them as micro categories.
Moreover, the *Reasoning* macro-category in specialized TE diagnostics include the following micro categories: *Apposition, Modifier Genitive, relative clause, elliptic expression, Meronymy, Metonymy, membership/representative, Quantity, Temporal, Spatial, common background, and general inferences*.

**Monotonicity**

Monotonicity was consistently included as a key phenomenon through all diagnostics datasets for decades. Monotonicity covers deduction, induction and no relation. Deductive reasoning (*downward-monotone*), a top-down approach, moves from a general premise to a specific hypothesis. For example, "all cats are beautiful" *entails* "my new white cat is beautiful" but the opposite is not true. On the other hand, inductive reasoning (*upward-monotone*) is the opposite of deductive reasoning. Inductive reasoning uses a bottom-up approach from specific premise to general hypothesis. For example, "I have a cat" *entails* "I have a pet", but the opposite is not true as I may have a rabbit or a tortoise. Moreover, the sentence "I have a cat" *has no relation with* "I have a rabbit" as both can be true simultaneously, this is a *non-monotony* example.

**World Knowledge & Common Sense**

FraCas and TE, knowing the omni-pervasive nature of world knowledge, did not explicitly categorize it, considering it as an area for future research. They acknowledged that world knowledge and common sense are required for various language understanding tasks, such as sense disambiguation, syntactic parsing, and anaphora resolution, and are therefore present to some degree in most examples. However, they also gathered examples where entailment relies not only on linguistic features, but also on knowing extra knowledge. On the other hand, ALUE and GLUE included *World Knowledge & Common Sense* in their diagnostics as a separate category. All diagnostics were minimizing assumptions of background knowledge by focusing on knowledge expected to be possessed by most people, regardless of cultural or educational background, such as common geographical, legal, political, technical, or cultural facts. For instance, the sentence "Paris is the capital of France" *contradicts with* "Grenoble is the capital of France", as we all know -by our world knowledge- that each country has only one capital. Also, we know -by our common sense- that if the first sentence is true, then the second will be false, because one city is one unique city and cannot be two cities. On the other hand, the sentence "Paris is a city in France" has *neutral relation with* the sentence "Grenoble is a city in France", as we all know that a country has many cities. And using our common sense, we know that if one is true then the other can also be true, as we have many cities, not a unique one.

Similarly, common sense was not explicitly categorized in FraCas and TE, but it was incorporated by ALUE and GLUE. An example of common sense is the contradiction relation of the sentence "everyone was shocked that …" with the sentence "everyone expected that …" because we cannot be shocked by something expected.

**Quantifiers**

Quantifiers were consistently included as a key phenomenon through all diagnostics datasets for decades. These datasets explore various quantifiers, including *universal* quantifiers (e.g., 'all'), *existential* quantifiers (e.g., 'exist one'), and other quantifiers like 'most', 'some', 'many', …. The core principle is that a broader quantifier in the premise generally entails a narrower quantifier in the hypothesis, but the reverse is not necessarily true. For example, "all students did the exam"

*entails* that "Mariam did the exam", but "Mariam did the exam" does *not entail* that "all students did the exam." Similarly, "most students passed the exam" *entails* "some students passed the exam", but "some students passed the exam" does *not entail* "most students passed the exam." Furthermore, "a student failed the course" does *not entail* "all students failed the course", although "all students failed the course" does *entail* "a student failed the course". The design of ALUE, GLUE, and TE specialized, included quantifiers as components of broader categories like *Reasoning* or *Logic*, while FraCas considered *quantifiers* as a separate macro category.

**Discourse**

Discourse focuses on the properties of the text as a whole that convey meaning by making connections between component sentences. Several types of discourse processing can occur at this level. Two of the most common types are anaphora resolution and discourse/text structure recognition. While all diagnostics include discourse tasks such as, *Anaphora*, but they do not consider it as a micro category of *Discourse*.
Discourse was only introduced in Specialized TE dataset hierarchy as a macro-category that includes *Coreference, Apposition, Anaphora zero, Ellipsis, and Statements*.

**Anaphora**

Anaphora, where a later expression refers back to an earlier one, is a controversial linguistic phenomenon in all diagnostics datasets. Anaphora affects the whole meaning as data consists of sentences taken out of context. These sentences often use referring expressions with anaphora that have no clear antecedent. Most diagnostics assumed that, where possible, expressions in the premise and hypothesis are co-referent.

For a noun, if there are definite noun phrases, each contains the "*squirrel*" and the "*dog*", then it is assumed that they are referring respectively to the same *"squirrel"* and *"dog"*. However, it is surprising how easily Anaphora can lead to tricky complications quickly: Does the sentence "the dog is brown" contradict "the animal is white" sentence? It depends on whether you want to force coreference between the dog and the animal, as the described situation may conceivably involve multiple creatures.

As for verbs, this can get more difficult with event coreference: clearly "John likes Gary" contradicts "John dislikes Gary", because they both describe a state that holds at the present time. But how about "John gave Gary a dollar" and "Gary gave John a dollar" sentences? They are contradictory if you force coreference of the giving event, but not if you allow that they happened in rapid sequence. However, if you allow them to happen in sequence, perhaps "John gave Gary a dollar reluctantly" should not contradict "John gave Gary a dollar enthusiastically". So where is the bright-line? It should be flexible.

GLUE solution as stated in GLUE, "to determine coreference, we judge based on the amount of corroborating information to a coreference judgment. For a noun, the same noun phrase description is enough corroborating information to induce coreference. But for a verb, its core arguments (e.g., John, Gary, and a dollar) must be co-referent in order to induce the coreference. The exception is if there is extra corroborating information (for example, a temporal modifier or quantifier like only) that uniquely establishes the event denoted by the given verb. (This will often happen in our examples that deliberately permute the arguments of verbs.)". GLUE stated that "This is not a perfect solution, and you may not agree with all of the judgments. If you wish to do your own evaluation without worrying about these coreference problems, there is an easy solution: collapse neutral and contradiction into a single non-entailment class. This works because the tension with coreference is between forcing an inadmissible coreference (i.e., one that leads to a contradiction) and allowing reference to a new entity (which almost always leads to a neutral)".

As for FraCas, Anaphora is one of its controversial aspects as stated in [91], "another, a more controversial aspect of FraCaS, is that it that the semantic relations it postulates between premises and hypotheses are only based on the semantics of the particular construction and the lexical meaning of the words involved. The data set contains no examples where the answers would depend on augmenting the premise with some background knowledge about the world. For example, it was assumed that the antecedent of an anaphoric expression such as he, it, and she was included in the premise, the pronoun was supposed not to be deontic or refer to something salient in the context that was not explicitly mentioned".

Despite these constraints, FraCaS includes several types of Anaphora, such as **Nominal Anaphora, Plural Anaphora, Inter-Sentential, E-type and Donkey Pronouns**. An example from FraCaS to illustrate this: Given the statement "Mary used her workstation," the questions "Is Mary female?", "Was Mary's workstation used?", and "Does Mary have a workstation?" are all answered "Yes" based *solely* on the information within the statement. In contrast, GLUE and ALUE include examples where the referent of an anaphor changes *depending on the context*. If the anaphor is correctly linked to its intended referent, the premise implies the hypothesis. Otherwise, the relation is a contradiction.

## 5. Conclusion

Natural Language Understanding (NLU) is a core component in Natural Language Processing (NLP). NLU capabilities evaluation has become a hot research topic in the last few years, resulting in the development of numerous benchmarks. These benchmarks, including diverse tasks and datasets, help do comparative evaluation of pretrained models via public leaderboards. Particularly, several benchmarks contain diagnostics datasets designed for investigation and fine-grained error analysis across a wide range of linguistic phenomena. This survey provides a review of existing NLU benchmarks, with a particular emphasis on their diagnostics datasets and the linguistic phenomena they covered. We present a detailed comparison and analysis of these benchmarks, highlighting their strengths and limitations in evaluating NLU tasks and providing in-depth error analysis. When highlighting the gaps in SoTA, we noted that there is no naming convention for macro and micro categories or even a standard set of linguistic phenomena that should be covered. For that reason, we formulated a research question regarding the evaluation metrics of the evaluation diagnostics benchmarks: "why do not we have an evaluation standard for the NLU evaluation diagnostics benchmarks?" similar to *ISO standard* in industry. We suggested to build a global hierarchy for linguistic phenomena under supervision of linguistics experts. We think that having evaluation metrics for evaluation diagnostics could be valuable to gain more insights when comparing the results of the same model on different diagnostics benchmarks.

**ABBREVIATIONS**

**NLI:** Natural Language Inference.

**NLU:** Natural Language Understanding.

**GLUE**: General Language Understanding Evaluation.

**ALUE**: Arabic Language Understanding Evaluation.

**FRACAS:** Framework for Computational Semantics.